\documentclass[11pt]{article}

\usepackage[preprint]{acl}

\usepackage{kotex}
\usepackage{multirow}
\usepackage{array}
\usepackage{subcaption}
\usepackage{tikz}
\usetikzlibrary{arrows.meta, positioning, fit, shapes}
\usepackage{graphicx}
\usepackage{booktabs}
\usepackage{siunitx}
\usepackage[export]{adjustbox} 
\usepackage{caption}
\usepackage{subcaption}
\usepackage{amsfonts}
\usepackage{amsmath}
\DeclareMathOperator*{\argmin}{arg\,min}

\usepackage{times}
\usepackage{latexsym}

\usepackage[T1]{fontenc}

\usepackage[utf8]{inputenc}

\usepackage{microtype}

\usepackage{inconsolata}

\usepackage{graphicx}

\title{Discovering Lexical Gaps Using Embeddings from Multilingual LLMs}

\author{
 \textbf{Yoonwon Jung},
 \textbf{Aaron S. Cohen},
 \textbf{Benjamin K. Bergen}
\\
\\
 Department of Cognitive Science, University of California San Diego
\\
\texttt{\{y5jung,aac011,bkbergen\}@ucsd.edu}
}

\begin{document}
\maketitle
\begin{abstract}

Lexical gaps are words that do not exist in certain languages. They pose challenges for building multilingual lexical resources, for machine translation, and for cross-lingual transfer. Existing lexical gap detection relies on human judgments or fixed conceptual taxonomies. We propose a data-driven framework for identifying cross-lingual lexical gaps. We extracted contextualized embeddings from Korean-English bilingual LLMs for Korean-to-English and English-to-Korean translation pairs. Combinations of LLMs, embedding types, dimensionality, and orthogonal transformations across 100 train-test splits yielded 4000 distinct embedding spaces in each source language. In each space, we computed the semantic similarity between each source word and its nearest neighbor in the target language, and compared their distribution for gap words versus non-gap words. In 94\% (Korean-to-English) and 97\% (English-to-Korean) of embedding spaces, gap words showed weaker cross-lingual semantic alignment than non-gap words. Logistic classifiers trained on unaligned embedding spaces can reliably separate gap words from non-gap words, achieving AUCs of 0.81 (Korean-to-English) and 0.76 (English-to-Korean) and retrieving 18/19 Korean and 26/27 English gap words. This approach provides a language-agnostic and taxonomy-free method for scalable lexical gap identification. 

\end{abstract}

\section{Introduction}

Human languages express concepts in various ways: as single words, collocations, or as ad hoc combinations of words. The availability of a conventional word to express a concept provides an efficient means of communication, achieving clarity with minimal effort \cite{lupyan2012words,rissman2023gaps,zipf2016human}. However, lexicalization differs across languages, resulting in occasional yet salient \textit{lexical gaps}, in which a concept encoded with a single lexical item in one language lacks a single-word equivalent in another language. Well-known examples include German \textit{Schadenfreude} `joy at another's misfortune' or Japanese \textit{komorebi} `sunlight filtering through trees.'

Lexical gaps offer a window into how different language communities carve up conceptual space. Prior work has suggested that lexicalization patterns across languages reveal variation in how humans think about and experience the world \cite{lomas2018experiential,lupyan2012words,rissman2023gaps,winawer2007russian}. To identify lexical gaps, therefore, is to discover cross-cultural conceptual differences (\citealp{wierzbicka1999emotions}).

Identifying lexical gaps is also useful as they pose practical challenges for machine translation (MT). Translation accuracy systematically suffers for sentences including culture-specific terms or expressions without literal translations in other languages \cite{bentivogli2000coping, khishigsuren2022using, yao2024benchmarking}. They also affect downstream performance for multilingual large language models (MLLMs), where they decrease performance on tasks such as natural language understanding and question answering \cite{ebrahimi2024zero}. This is especially problematic for cultural appropriateness, as many lexical gaps denote culture-specific concepts that models fail to capture. In short, accurately identifying lexical gaps is critical not only for comparative linguistic and psychological research, but also for improving the performance of MLLMs.

Despite their importance, lexical gaps remain under-identified. Most existing multilingual lexical resources do not explicitly mark lexical gaps, as language-specific words are often omitted or mapped to approximate translations (e.g., \citealp{bond2013linking, navigli2012babelnet}). Moreover, lexical gaps are difficult to identify. Detecting lexical gaps typically requires substantial manual annotation by trained native bilinguals \cite{bentivogli2000coping, janssen2004multilingual}.

Recent approaches to this problem adopt top-down methods that rely on fixed conceptual taxonomies to detect lexical gaps. Although this approach works well for the kinship domain \cite{khalilia2023lexical, khishigsuren2022using}, it presupposes that the target domain's conceptual hierarchies are well-defined and agreed upon, and that the vocabulary size is manageable for manual classification based on the conceptual taxonomy. These assumptions limit the applicability of the same approach to domains such as emotion, cognition, or values, with larger vocabulary sizes and unclear conceptual hierarchies.

We propose a data-driven, bottom-up approach to identify cross-lingual lexical gaps that does not rely on predefined conceptual taxonomies. It operates directly on existing word lists and uses distributional semantic measures from the embeddings of bilingual LLMs to quantify the \textit{gappiness} of individual words across a pair of languages. We evaluate the approach in the domain of emotion, leveraging a high-quality, hand-crafted dataset of Korean and English emotion words that explicitly identifies lexical gaps (EVOKE; \citealp{jung2026evoke}). We demonstrate that this method successfully distinguishes gap words (words without a single-word translation in the other language) from non-gap words and quantifies the degrees of \textit{gappiness}. Moreover, a comparative analysis of alignment approaches shows that embeddings from MLLMs can be used without additional cross-lingual alignment. This approach provides a principled way to narrow the search space for lexical gaps in large multilingual resources, reducing the reliance on native experts and enabling systematic analysis of lexical gaps for both linguistic research and multilingual model evaluation.

\section{Related Work}

\subsection{Definition of lexical gaps}
The term \textit{lexical gap} broadly refers to the absence of a lexical unit describing a concept in a natural language (for an overview, see \citealp{ivir1977lexical}). While some arise at the morpho-syntactic level (e.g. \citealp{bentivogli2000coping}), we use the term to refer to cross-lingual semantic gaps, as semantic-level lexical gaps reveal unique dimensions of experience that one language captures conventionally but that are absent in other languages \cite{ivir1977lexical, lomas2018experiential, wierzbicka1999emotions}. 

More specifically, we define a lexical gap as a concept lacking a single-word equivalent in the target language, even if it can be expressed via collocations or ad-hoc word combinations \citep{janssen2004multilingual, li2024translation, wierzbicka1997understanding}. Although some prior work adopts a broader definition (e.g. \citealp{bentivogli2000coping, khishigsuren2022using}) that includes conventional multi-word collocations, we adopt a stricter approach. This narrower, word-specific definition avoids the need for subjective judgments about which multi-word phrases are conventional, and facilitates more comparable evaluation across languages.


\subsection{Lexical gaps and MLLM performance}
A growing body of work demonstrates that lexical gaps pose challenges for multilingual NLP systems. In the context of MT, \citet{khishigsuren2022using} showed that MT systems mistranslate kinship terms that have lexical gaps across languages: sentences containing kinship gaps (e.g., \textit{brother} in English, which lacks a direct equivalent in many languages) produce significantly larger semantic mismatches than sentences without lexical gaps across five target languages (Russian, Korean, Japanese, Hungarian, and Mongolian). 

Lexical gaps also affect downstream task performance beyond MT. \citet{ebrahimi2024zero} showed that lexical gaps negatively affect multilingual question-answering (QA). MT of QA tasks involving lexical gaps in English into multiple languages (Catalan, German, Farsi, Hindi, and Vietnamese) leads to worse performance than when using oracle translations. Zero-shot QA performance in target languages was worse when the task involved lexical gaps in English, compared to when the task was translated to English using oracle translation. These trends increased with typological distance. Furthermore, lexical gaps can undermine the cultural appropriateness of MLLM-generated empathetic responses. \citet{lee2025multilingual} show that English-centric MLLMs struggle to respond appropriately to scenarios involving Korean emotion words without lexical equivalents in English.

In summary, lexical gaps remain a systematic source of error and performance degradation for MLLMs, particularly in languages that are typologically distant from English, and lexical gaps can undermine not only semantic accuracy but also generating culturally appropriate and empathetic MLLM responses. This underscores the need for principled approaches to identify lexical gaps, to improve model training and evaluation.

\subsection{Existing resources on lexical gaps}

Lexical gaps have long been recognized as a challenge for constructing multilingual lexical resources \cite{bentivogli2000coping, bond2020some, khishigsuren2022using, ordan2007hebrew}. Few existing resources explicitly identify them, as language-specific word meanings are often excluded from datasets or are mapped to loose equivalents, as in Open Multilingual WordNet \cite{bond2013linking} and BabelNet \cite{navigli2012babelnet}. While such approximations facilitate broad coverage, they obscure cases where no true single-word equivalent exists. 

Several approaches have been adopted to address this limitation. Extensions on MultiWordNet (MWN) \cite{pianta2002multiwordnet} indexed lexical gaps (e.g., in Hebrew and Italian) by leveraging manual classification by bilingual coders and decision trees \cite{bentivogli2000coping, ordan2007hebrew}. However, these approaches are dependent on language-specific rules for constructing decision trees and on native bilinguals’ manual classification efforts. More recently, several studies have curated a dataset of lexical gaps in the Kinship domain through semi-automated approaches. These studies leveraged conceptual hierarchies among kinship concepts formalized by existing theory and hyponym relationships \cite{khalilia2023lexical, khishigsuren2022using, li2024translation}. While effective, these approaches lack generalizability to other conceptual domains where no such objective conceptual hierarchy exists (such as emotions), or where the vocabulary size in that domain for a given language is more extensive. 

\subsection{Lexical Gaps in Emotion Terms}
The semantic domain of emotions presents a particularly compelling test case for these challenges. Unlike kinship, emotion concepts do not belong to a widely agreed-upon, theory-neutral conceptual hierarchy. Competing theories disagree on whether emotions can be organized into first-order and higher-order categories, which emotions count as basic, and whether such distinctions are culturally universal \cite{barrett2006emotions, ekman1992argument, johnson1989language, plutchik2001nature, reisenzein1995oatley, russell1994fuzzy, smith2009critiquing}.\footnote{For example, while kinship terms expressing concepts like \textit{paternal aunt} and \textit{maternal aunt} can be straightforwardly organized under a higher-order concept \textit{aunt}, hierarchies for emotion concepts are far less clear. Whether emotions such as \textit{depressed} or \textit{devastated} should be treated as subtypes of \textit{sad}, or whether \textit{sad} itself constitutes a basic emotion, remains theoretically contested.} As a result, top-down approaches that rely on fixed conceptual taxonomies are difficult to apply to emotion, like other domains, in a principled and scalable manner across languages. 

Moreover, emotion words exhibit substantial cross-linguistic variation. Emotion word semantics align less across languages than do words in other domains \cite{jackson2019emotion, thompson2020cultural}, and various studies have reported language-specific emotion words that are untranslatable to other languages \cite{mesquita2016cultural, schmidt1999korean, wierzbicka1992defining, wierzbicka1999emotions}.

In what follows, we develop and test a method to automatically detect lexical gaps, using the domain of emotion. To do this requires a high-quality, hand-curated bilingual dataset of emotion words that explicitly identifies lexical gaps, providing a combination of coverage, annotation quality, and explicit gap marking. We use the EVOKE dataset, focusing on English and Korean \citep{jung2026evoke}. This dataset enables examination of lexical gaps across typologically distinct languages. We demonstrate a scalable and domain-general lexical gap detection method using words in typologically distinct languages and in a semantically challenging domain.

\section{Materials and Methods}

\begin{figure*}[t!]
    \centering
    \includegraphics[width=\textwidth]{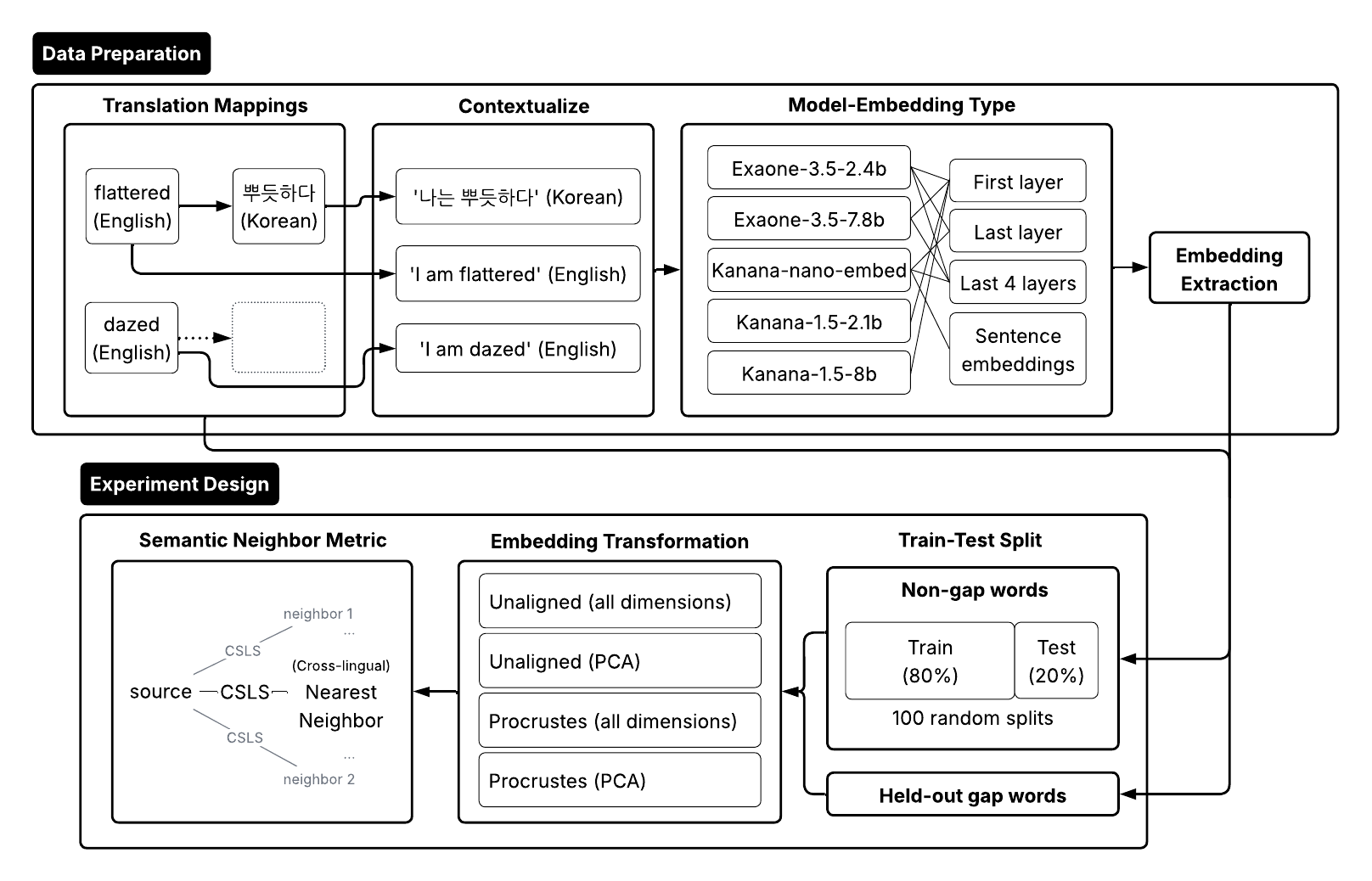}
    \caption{Overview of the Experimental Pipeline.}
    \label{fig_flowchart}
\end{figure*}

We search for lexical gaps by examining cross-lingual semantic neighbors in embedding spaces of multilingual language models. Semantic neighbors have previously been used as a measure of cross-lingual semantic alignment \cite{karidi2024locally, peng2024concept, thompson2020cultural}. However, point-to-point translation retrieval from the embedding spaces is less feasible, as precision at 1 (p@1) is reported to be generally low in both aligned and unaligned multilingual embeddings, especially for unseen test words and for typologically distinct language pairs \cite{peng2024concept}.\footnote{See also Appendix~\ref{sec:appendix_p@1} for similar p@1 results.} Moreover, p@1 is not applicable in the presence of lexical gaps, since no ground-truth translation exists for gap words.

Therefore, we extend the intuition as follows: if a source language word has a single-word equivalent in the target language, the nearest neighbor in the target language should exhibit stronger similarity to the source word than those of lexical gaps, for which no single-word equivalent exists. We operationalize this contrast using a nearest-neighbor–based metric and evaluate how different embedding spaces defined by model-embedding combinations and embedding transformation choices distinguish gap words from non-gap words. An overview of the experimental pipeline is in Figure~\ref{fig_flowchart}.

\subsection{Word Data}
As mentioned above, reliable evaluation of lexical gaps requires a dataset that explicitly distinguishes gap and non-gap words within a well-defined semantic domain. We therefore focus on Korean and English emotion adjectives with and without translations in the other language.\footnote{Adding non-emotion adjectives to augment the training source improved performance in some embedding configurations for Korean, but yielded substantial performance drop for English. See Appendix~\ref{sec:appendix_non+emotion} and Figure~\ref{fig_appendix_medianauc_more_train} for details on obtaining embeddings for non-emotion words and full experiment results.} Korean and English emotion adjectives were obtained from the EVOKE dataset, which provides comprehensive coverage of emotion words and lexical gaps in each language \cite{jung2026evoke}.

EVOKE aggregates emotion words from multiple existing resources and annotates them according to criteria designed to capture their properties as emotion terms. These criteria include human acceptability judgments in diagnostic sentence frames such as ``I am X'' and ``I feel X''. We included only those emotion words that were judged acceptable in both diagnostic sentence frames (``I am X'' and ``I feel X'') to ensure that all selected words function naturally as emotion predicates and to allow the extraction of contextualized embeddings of emotion words using a shared linguistic context (``I am X'').

All words were restricted to adjectives for several reasons. First, EVOKE provides acceptability annotations primarily for adjectives and verbs. Second, adjectives exhibit greater lexical diversity than verbs in both Korean and English in the emotion domain, resulting in a larger and more balanced dataset. For Korean, we utilized adjectives in the inflectional form that fit the sentence frame, provided in EVOKE, to ensure syntactic compatibility.

\subsection{MLLMs and Embeddings}
For the semantic neighbor-based approach to be effective, it is crucial to select models and embedding representations that reliably encode lexical semantics in both Korean and English. Therefore, we adopt a multi-stage strategy to select MLLMs and embedding types. 

\subsubsection{Model Candidates}
\label{sec:model_candidate}
We considered multiple Korean–English bilingual LLMs drawn from three model families: the EXAONE \cite{an2024exaone}, Kanana \cite{bak2025kanana}, and HyperCLOVA X \cite{yoo2024hyperclova} model families. To minimize interference from languages other than Korean and English and to improve the interpretability of embedding representations, we adopted five model selection criteria: (1) trained exclusively on Korean and English; (2) exclusion of models based on mixture-of-experts architectures; (3) exclusion of reasoning models; (4) preference for base models over instruction-tuned variants when available; and (5) selection of the most recent model unless architectural differences justified an earlier version. 

\subsubsection{Embedding Extraction}
To extract comparable lexical representations across languages, we embedded each emotion adjective in a single minimal sentence context (``I am X” for English; ``나는 X” for Korean, where the target word was substituted for X). This neutral framing captures a prototypical use of emotion predicates while minimizing contextual influence and variability \cite{clore1987psychological, havaldar2023multilingual, jung2026evoke}. Acceptability of emotion words in the sentence frame was ensured using human judgments provided in the EVOKE dataset. 

For words tokenized into multiple subword units, the corresponding token embeddings were averaged. For each model, we extracted multiple embedding types: from the first hidden layer, the last hidden layer, the average of the last four hidden layers, and sentence embeddings (when available).\footnote{\texttt{kanana-nano-embedding} is the embedding model built upon \texttt{kanana-nano-base}. Therefore, for \texttt{kanana-nano} model, sentence embeddings were obtained using the embedding model, while the backbone base model was used to extract token-level embeddings.} Sentence embeddings were included because a fixed sentence frame was used to extract embeddings for all target words.

\subsubsection{Translation Mappings}
For a translation gold standard, Korean and English non-gap emotion word embeddings were paired based on the translation mappings from EVOKE. This resource includes many-to-many translational mappings from source words to target words. We retained only translation pairs in which the source word maps to only one target language word (one-to-one or many-to-one), as having multiple target words complicates the definition of the gold standard for model training. As a result, 209 Korean-to-English pairs (209 unique Korean, 157 unique English words) and 235 English-to-Korean pairs (235 unique English, 175 unique Korean words) were retained. 19 Korean gap words and 27 English gap words were left without translation mappings.

\subsubsection{Model-Embedding Combinations Selection Results}
\label{sec:top10}
We next identify model–embedding combinations that reliably capture translational equivalence. For each candidate language model and embedding type, we compared cosine similarities of translationally equivalent non-gap word pairs (matched pairs; e.g., \textit{sad}–\textit{슬프다} ‘sad’) and non-equivalent pairs (mismatched pairs; e.g., \textit{sad}–\textit{행복하다} ‘happy’). For each model and embedding type, a t-test on the difference between the mean cosine similarities of matched and mismatched pairs was performed, along with Cohen’s \textit{d} as an effect size.

\begin{table}[!t]
\nolinenumbers
\centering
\begin{tabular}{
  @{}>{\setlength{\parindent}{0pt}}p{0.5\columnwidth}
  p{0.41\columnwidth}
}

\hline
Model & Embedding Type \\ \hline

{\small\texttt{Exaone-3.5-2.4b-instruct}} & First layer \\
& Last layer \\
& Last 4 layers \\

{\small\texttt{Exaone-3.5-7.8b-instruct}} & First layer \\
& Last 4 layers \\

{\small\texttt{Kanana-nano}} & Last layer \\
& Last 4 layers\\
& Sentence embedding \\

{\small\texttt{Kanana-1.5-2.1b-base}} & First layer \\
{\small\texttt{Kanana-1.5-8b-base}} & First layer \\

\hline
\end{tabular}
\caption{Top 10 model-embedding combinations retained for downstream experiments and evaluation. `Last 4 layers' refers to the averaged embedding of the last four layers.}
\label{tab_top10}
\end{table}

The 10 combinations of embedding type and language model that maximized the separation between matched and mismatched Korean–English emotion word pairs (as measured by effect size) were retained (see Table~\ref{tab_top10}). Two language models from the EXAONE family and three language models from the Kanana family were retained, with one to three embedding types from each model. No embeddings from the HyperCLOVA X model ranked in the top 10. The full results are reported in Appendix~\ref{sec:appendix_top10}. This selection step ensures that subsequent gap detection is evaluated only on embedding spaces that encode semantic correspondence across Korean and English words well.

\subsection{Train-test Split}
For the remaining feature selection and downstream evaluation, we randomly split non-gap words in each language into train (80\%) and test (20\%) sets. This train-test split was repeated across 100 random seeds, yielding different train and test sets for each random seed. Summary statistics of test words sampled across random seeds are reported in Appendix~\ref{sec:appendix_traintest}.

\subsection{Embedding Transformation}
\label{sec:embed_transformation}
We extract features for 4 different embedding transformations (described below) using the 10 selected model-embedding combinations. Features are extracted separately for each source language.

Following previous work on cross-lingual embedding alignment \cite{huertas2023exploring,schonemann1966generalized, peng2024concept}, we consider dimensionality reduction using Principal Component Analysis (PCA), orthogonal alignment using Procrustes analysis, and their combination, alongside untransformed embeddings. We thus projected all source words through 4 embedding transformations per training condition: (1) unaligned (all dimensions), (2) unaligned (PCA), (3) Procrustes (all dimensions), and (4) Procrustes (PCA). 

For each model-embedding combination (n=10), we used the combined source and target word embeddings in the train set for PCA, and the mappings of source-to-target words and their embeddings in the train set for Procrustes analysis. PCA was applied to reduce the embeddings to 256 dimensions, using only the words in the training set.\footnote{The number of dimensions for PCA was selected to retain sufficient dimensions without exceeding the size of the training vocabulary. See Appendix~\ref{sec:appendix_pca_dim} for further analysis of different PCA dimensionalities.} Procrustes analysis was performed to learn an orthogonal transformation matrix that aligns source word embeddings to target word embeddings \cite{schonemann1966generalized, peng2024concept}. Procrustes orthogonal transformation matrix $W^*$ is derived through Singular Value Decomposition (SVD) of $X^TY$, where $U\Sigma V^T = \mathrm{SVD}(X^TY)$. $X \in \mathbb{R}^{n \times d}$ is the matrix of source language embeddings, and $Y \in \mathbb{R}^{n \times d}$ is the matrix of target translation embeddings. The $n$th row of $X$ and $Y$ refer to the $n$th translation pair of source to target language.
\begin{equation}
    W^* = \argmin_{W \in O_d(\mathbb{R})} \|XW - Y\|_F = UV^T
\end{equation} 

We mean-center all source and target words using means of training words in each language with L2 normalization for all 4 combinations.

\subsection{Semantic Neighbor Metric}
For each source word in the test set and the held-out gap word pool, we calculated its similarity to the nearest neighbor in the target language. We focus on similarity scores to capture graded semantic correspondence. Cosine similarity with cross-domain local scaling (CSLS) was used as a similarity metric \cite{conneau2017word}, which scales raw cosine similarity by the hubness of nearby source and target words. We set the number of neighbors to 10 for CSLS calculation \cite{conneau2017word}. We adopt CSLS throughout the main analyses, as prior work has shown it to outperform raw cosine similarity in cross-lingual semantic alignment tasks \cite{conneau2017word, peng2024concept}. (Our experiments on p@1 also demonstrated improved p@1 under CSLS; see Appendix~\ref{sec:appendix_p@1}).

On the resulting 4000 embedding spaces (100 random seeds for train-test splits × 10 model-embedding combinations × 4 embedding transformation combinations), we computed the CSLS score between each source word (both gap and non-gap test words) and its nearest neighbor in the target language. All gap words were evaluated in each train-test split (across 100 random seeds), while the number of observations of non-gap test words varied, as different sets of words were sampled for each random seed (see Appendix~\ref{sec:appendix_traintest}). 

\section{Evaluation}
Based on the proposed metric values calculated from 4000 different embedding spaces, we employ two approaches to quantify the \textit{gappiness} of words and evaluate the separation between gap and non-gap words (gap word separability).

\subsection{Gap vs Non-gap Test word AUC}
\subsubsection{Evaluation Setup}
We compare the distribution of the nearest neighbor CSLS metric between gap words and test words across all 4000 unique embedding spaces, and use the Area Under the Curve (AUC) of each embedding space. The resulting AUC measures the degree of separation of gap words from non-gap words based on CSLS scores. Higher AUC indicates greater separability between the two distributions (CSLS score distributions of gap words and non-gap test words), with gap words generally exhibiting lower CSLS scores than non-gap test words across the embedding spaces. Using this AUC, we confirm our operationalization of gap words based on the nearest neighbor in embedding space by looking at the distribution of AUCs across all embedding spaces. 

Moreover, we take the median AUCs across all 100 train-test splits for each of the 40 embedding space configurations (10 model-embedding combinations × 4 embedding transformations). Using these median AUCs, we first calculate Spearman's rank correlation of the AUC of Korean and English as source languages to assess cross-linguistic consistency. We also assess which of the 40 embedding spaces yield the highest median AUC, and further identify a subset of embedding spaces that systematically outperform others across languages. These findings were used to guide feature selection for logistic classifiers.

\subsubsection{Evaluation Results}
\label{sec:eval_result_1}
\begin{figure}[t!]
    \centering
    \includegraphics[width=0.98\columnwidth]{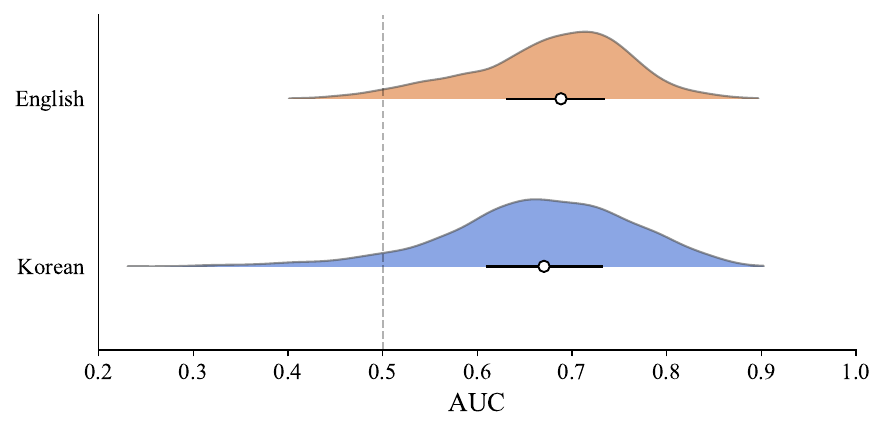}
    \caption{Distribution of AUCs across all embedding spaces and random seeds for each source language. Bars represent inter-quartile ranges, and middle points indicate median AUCs for each source language.}
    \label{fig_auc}
\end{figure}

The distribution of AUCs in each embedding space per source language is in Figure~\ref{fig_auc}. 94\% of the embedding spaces in Korean and 97\% in English exhibited AUCs above 50\%. This confirms the operationalization of gap words across most of the embedding spaces: gap words have lower CSLS similarity to their nearest neighbor than non-gap test words in most of the embedding spaces. The correlation between Korean and English AUC scores was \textit{r} = 0.58 (Spearman's rank correlation), showing moderate consistency of gap word separability across the two source languages.

Comparison of individual AUC scores revealed the embedding spaces with the best gap separation as the last layer of the \texttt{kanana-nano} model (unaligned, without PCA) for Korean, and the first layer of the \texttt{Exaone-3.5-2.4b-instruct} model (unaligned, without PCA) for English (see Figure~\ref{fig_appendix_medianauc}). Unaligned embeddings generally outperformed aligned spaces across both languages, although the best model–embedding types differed between Korean and English.

\subsection{Logistic Classifiers}
\subsubsection{Evaluation Setup}
We trained logistic regression classifiers to estimate the probability of a word being a lexical gap (\textit{gappiness}) using Leave-One-Out Cross-Validation (LOOCV). Predictor variables consisted of CSLS scores from each embedding space configuration, computed as the median CSLS score between each source word and its cross-lingual nearest neighbor across random seeds.\footnote{Because different sets of test words were selected across 100 random seeds, the number of non-gap test words substantially exceeded the number of gap words (see Appendix~\ref{sec:appendix_traintest}). Therefore, we used balanced class weights.} 

Based on the results in Section~\ref{sec:eval_result_1}, two sets of models were compared: the reduced models used CSLS scores from 20 embedding space configurations (10 model-embedding combinations × 2 embedding transformation combinations), while the full models used CSLS scores from 40 embedding space configurations (10 model-embedding combinations × 4 embedding transformation combinations). To address the large degree of correlation among features, we employed an L1 penalty with varying regularization strengths. 

Then, we propose a probability threshold for gap word retrieval based on the experiments on precision and recall over varying thresholds. This threshold could also be used for efficient manual lexical gap detection from other lexical resources. We also suggest some additional inspection criteria based on the characteristics of the gap words that the models struggle to identify.

\subsubsection{Evaluation Results}

\begin{figure*}[t!]
  \centering
  \begin{subfigure}{\textwidth}
      \centering
      \includegraphics[width=0.98\textwidth]{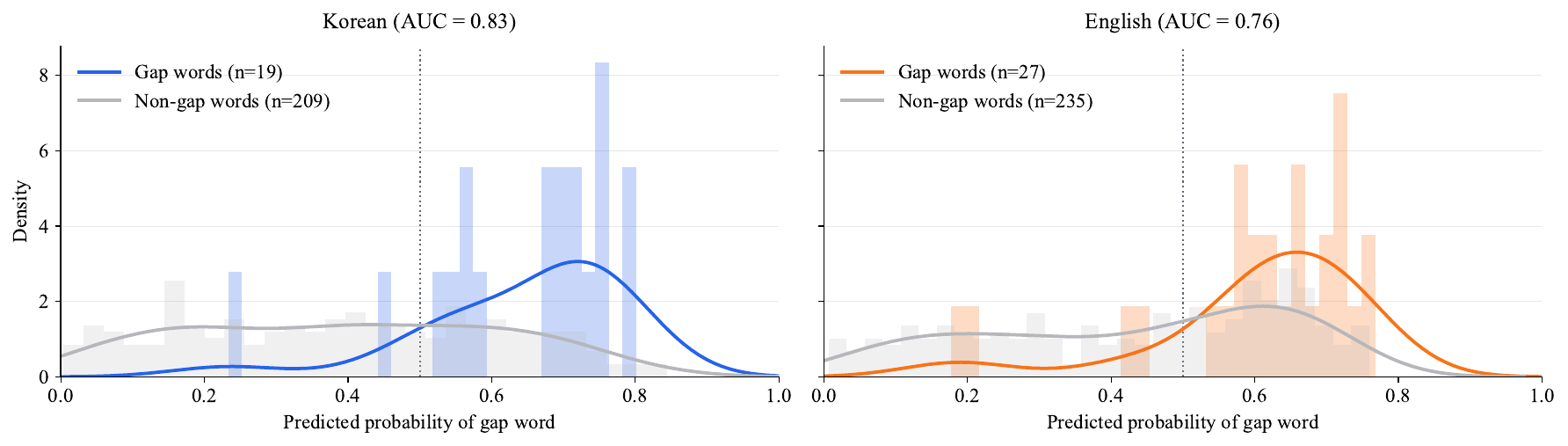}
      \caption{All embedding spaces as predictors (full models)}
      \label{fig_logistic_all}
  \end{subfigure}
  \hfill
  \begin{subfigure}{\textwidth}
      \centering
      \includegraphics[width=0.98\textwidth]{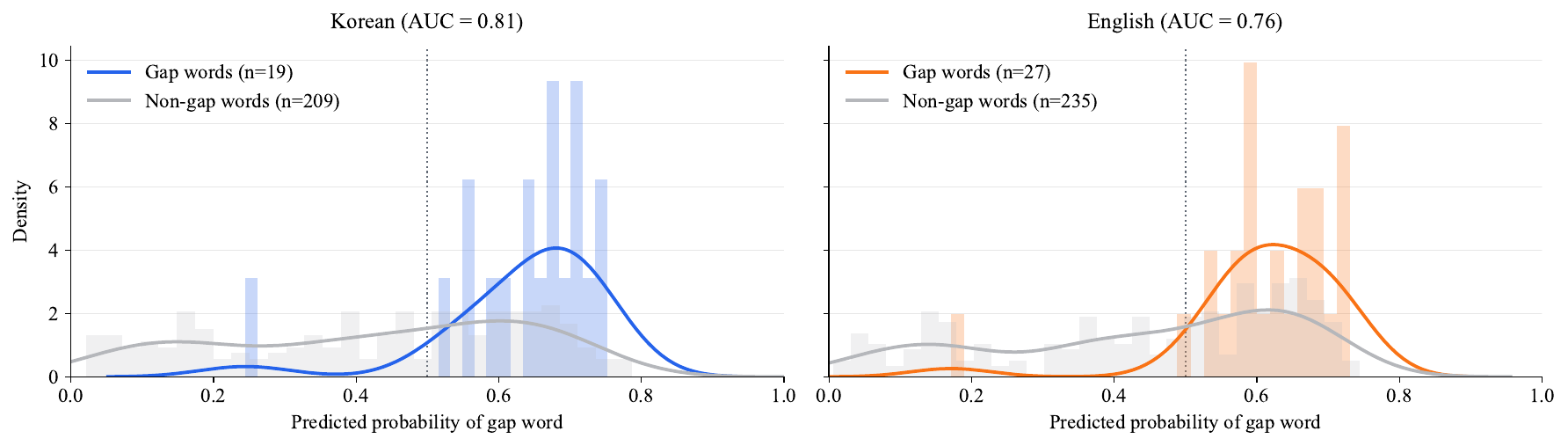}
      \caption{Unaligned embedding spaces as predictors (reduced models)}
      \label{fig_logistic_unaligned}
  \end{subfigure}
  \caption{Distribution of predicted probabilities of gap and non-gap words across predictor sets and source languages. Probabilities are estimated using logistic classifiers with an L1 penalty ($C$ = 0.1).}
  \label{fig_logistic}
\end{figure*}

\begin{figure*}[t!]
    \centering
    \includegraphics[width=0.98\textwidth]{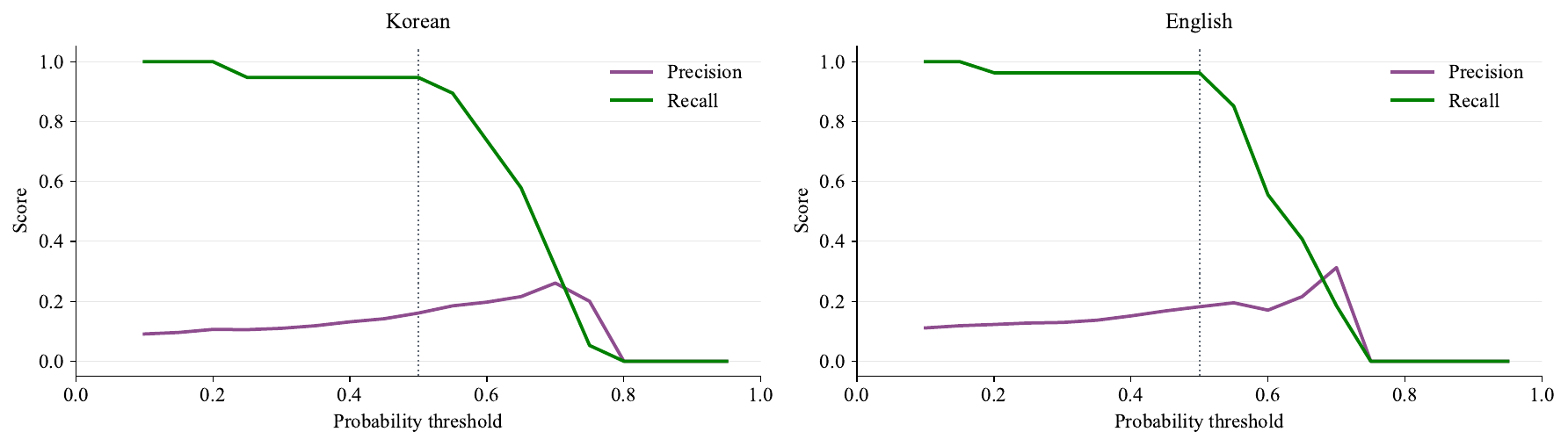}
    \caption{Precision and recall by different probabilities as thresholds. Probabilities are estimated using logistic classifiers trained on unaligned spaces with an L1 penalty ($C$ = 0.1).}
    \label{fig_precision_recall}
\end{figure*}

After experimenting with different regularization strengths (see Figure~\ref{fig:appendix_l1_penalty} in Appendix), we report results using $C$ = 0.1 for a balance between AUC and retrieval for both full and reduced models. The full models exhibited higher AUC (AUC = 0.83) than the reduced models (AUC = 0.81) for Korean source words, but the two models had the same AUCs (AUC = 0.76) for English.

We use a probability threshold of 0.5 to retrieve gap words (\textit{gappiness} threshold). This value corresponds to an above-chance probability and captures the majority of gap words in both languages. Experiments on precision and recall over varying thresholds also confirmed that the 0.5 threshold maintains near-maximal recall while retaining moderate precision across both languages (see Figure~\ref{fig_precision_recall}). The reduced models retrieved more words than the full models in both Korean (18/19 versus 17/19 retrieval) and English (26/27 versus 23/27 retrieval). Full results across different regularization strengths are reported in Figure~\ref{fig_logistic}.

The one Korean gap word below the 0.5 threshold is \textit{기고만장하다} (p = 0.245) in Korean, which is polysemous; one of the meanings has a gap in English but not the other.\footnote{It corresponds to both 1) `be enraged' and 2) `be in one’s glory' \cite{jung2026evoke}.} These multiple meanings could not be sufficiently captured when using a single context (``I am X” for English; ``나는 X” for Korean, where the target word was substituted for X) to extract contextualized embeddings. The English gap word below this threshold was \textit{lovesick} (p = 0.171). As a compound word, the token-based embeddings might have struggled to capture it (e.g., see \citealp{komiya2023composing}). 

\section{Discussion}
We have proposed a novel data-driven framework to identify cross-lingual lexical gaps using embeddings from MLLMs. Across different embedding spaces, gap words consistently exhibited decreased similarity to their cross-lingual nearest neighbor compared to non-gap words. The proposed method also identifies the embedding spaces, defined by model choice, embedding type, and embedding transformation, that most effectively separate gap from non-gap words.

Logistic classifiers trained exclusively on unaligned embedding spaces (reduced models) achieved AUCs of 0.81 for Korean and 0.76 for English, retrieving 18/19 Korean and 26/27 English gap words. This is comparable to the performance of classifiers trained on both unaligned and aligned spaces (full models).\footnote{This comparability was observed using 256 PCA dimensions. When PCA dimensionalities were selected separately for aligned and unaligned embedding spaces based on the best gap word separability, reduced models still achieved better retrieval but worse AUC than full models. See Appendix~\ref{sec:appendix_pca_dim} for full experiments on PCA dimensionalities.} Building on previous findings (e.g., \citealt{peng2024concept}), this result suggests that multilingual embedding spaces extracted from high-quality bilingual LLMs encode cross-lingual lexical information supporting the identification of lexical gaps without additional alignment.

Based on the gap probability estimated from the logistic classifiers, we propose using a probability threshold of 0.5 for lexical gap identification. Because the goal is to maximize recall, this threshold includes as many true gap words as possible while retaining moderate precision. Using this threshold for lexical gap identification in future work will substantially reduce the burden of manual inspection on native experts (see Section~\ref{sec:replication_application} and Appendix~\ref{sec:appendix_application} for detailed guides). Additional targeted inspection of polysemous or compound words could support more precise lexical gap identification, although it could vary by domain or language. 

Taken together, these findings demonstrate the feasibility of scalable lexical gap detection. The experimental setting (emotion domain, Korean-English language pair) constitutes a challenging test case due to the abstract and variable nature of emotion concepts, the lack of a well-defined conceptual taxonomy of emotions, and the typological distance between the two languages. As few existing resources explicitly identify lexical gaps \cite{bond2013linking,navigli2012babelnet}, and existing approaches for identifying lexical gaps lack generalizability to other conceptual domains \cite{khalilia2023lexical,khishigsuren2022using,li2024translation}, the proposed framework provides a novel way to be applied to any domain or language pair. Future work could utilize the proposed method to construct more comprehensive multilingual lexical resources that include explicit identification of lexical gaps. The identified lexical gaps could be used to test and improve machine translation performance, and facilitate cross-cultural research.

\section{Replication and Application}
\label{sec:replication_application}
The proposed framework is applicable across domains and language pairs. This enables conceptual replication and practical application of the method to other language pairs and datasets.\footnote{The code is publicly available at \url{https://github.com/UCSD-Language-and-Cognition-Lab/lexical_gap_embedding}.} However, there are several conditions to be satisfied.

First, to conceptually replicate the framework in other language pairs or domains, one should use MLLMs satisfying the criteria in Section~\ref{sec:model_candidate} to obtain embeddings. Using MLLMs trained exclusively on the two target languages is particularly important for a precise replication of the results, although other requirements could be relaxed based on model availability. In addition, a multilingual lexical resource meeting the following conditions is required as ground truth: (1) it should have comprehensive lexical coverage in each of the languages within the same domain, and (2) translation mappings should be precise and explicitly indicate lexical gaps. Lastly, since this framework adopts a data-driven approach, each language and domain must contain a sufficiently large vocabulary to estimate an accurate orthogonal transformation matrix for cross-lingual alignment \cite{peng2024concept}.

By contrast, applying this approach to identify new lexical gaps imposes fewer constraints on the choice of lexical resources. The strong performance of classifiers trained on unaligned embedding spaces indicates that the method can be used even without explicit translational mappings. Therefore, a multilingual resource or two separate monolingual lexical resources containing words in the same domain, with comprehensive lists of words in each language, is sufficient to build a multilingual lexical resource with explicit lexical gap identification. Moreover, to obtain the embeddings, using MLLMs trained exclusively on the two target languages provides the cleanest instantiation of our tested methodological pipeline. See Appendix~\ref{sec:replication_application} for a step-by-step guide.

\section*{Acknowledgments}
We would like to thank Samuel Taylor, Chaitanya Kapoor, and others who reviewed the code and provided suggestions for improvement. We also thank the reviewers for their valuable feedback and comments.

\bibliography{custom}

\appendix

\section{Limitations}
While this approach introduces a novel method for identifying cross-lingual lexical gaps, this study is limited to the emotion domain. In addition, this approach was evaluated only on Korean–English data. Future work could expand this approach to other domains and language pairs to confirm the generalizability of this method.

\section{Model-Embedding Selection}
\label{sec:appendix_top10}

\begin{figure*}[p]
    \centering
    \includegraphics[width=0.85\textwidth]{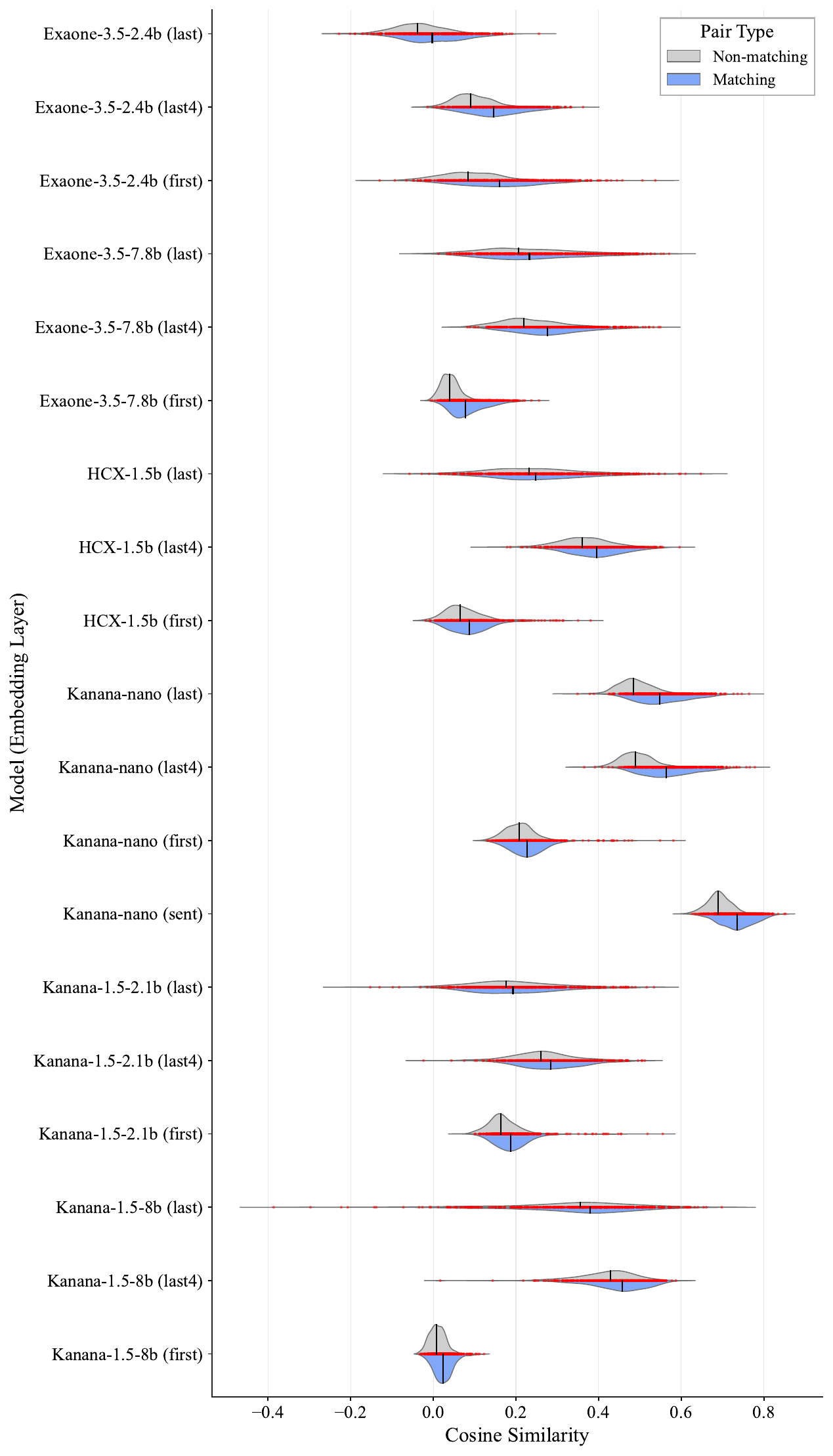}
    \caption{Distribution of cosine similarity across models and embedding layers. Red dots indicate data points in the `Matching' Pair Type condition.}
    \label{fig_appendix_fullcs}
\end{figure*}

We report the full lists of LLMs and their embedding variants, as well as the cosine similarity distribution of matched and mismatched pairs for each model-embedding combination in Figure~\ref{fig_appendix_fullcs}. Six models are tested across three layers (with additional sentence embeddings from \texttt{kanana-nano} model). We also report the test statistics of the differences in mean of the cosine similarity of matched and mismatched pairs and the effect sizes (Cohen's \textit{d}) in Table~\ref{tab_appendix-stats}. All combinations of model-embedding were significant after False Discovery Rate correction on p-values. We selected the 10 combinations that showed the highest effect size. 

\section{Descriptive Statistics of Train, Test, and Gap Data}
\label{sec:appendix_traintest}

Among 209 Korean-to-English pairs, 42 words were sampled for a test set, and 167 words were selected for a train set for each random seed. Among 235 English-to-Korean pairs, 47 words were sampled for a test set, and 188 words were selected for a train set for each random seed. The number of gap words remained the same across random seeds. Distribution of test word sampling across 100 random seeds is presented in Table~\ref{tab_appendix-test-sampling}.

\begin{table}[ht!]
\begin{tabular}{lcc}
\hline
\begin{tabular}[c]{@{}l@{}}The number of\\ random seeds\end{tabular} & English & Korean \\
\hline
8  & 1  & 0 \\
9  & 1  & 2 \\
10 & 0  & 0 \\
11 & 2  & 0 \\
12 & 2  & 2 \\
13 & 0  & 3 \\
14 & 9  & 6 \\
15 & 7  & 9 \\
16 & 17 & 15 \\
17 & 22 & 17 \\
18 & 19 & 16 \\
19 & 31 & 21 \\
20 & 22 & 31 \\
21 & 30 & 16 \\
22 & 13 & 11 \\
23 & 16 & 19 \\
24 & 14 & 16 \\
25 & 10 & 8 \\
26 & 7  & 4 \\
27 & 4  & 5 \\
28 & 6  & 6 \\
29 & 1  & 2 \\
30 & 0  & 0 \\
31 & 0  & 0 \\
32 & 1  & 0 \\
\hline
Total & 235 & 209 \\
\hline
\end{tabular}
\caption{Distribution of test word sampling across 100 random seeds. Each row shows how many random seeds resulted in a particular number of test words being selected for each language. For example, 1 English word was sampled across 8 different random seeds.}
\label{tab_appendix-test-sampling}
\end{table}

\section{Precision@1}
\label{sec:appendix_p@1} 

We first examined how the target word can be recovered through nearest neighbor metrics in the embedding space. We compute the average precision@1 (p@1) for non-gap words, and report on the results separately for Korean and English. P@1 is defined as the percent of total source language words where the nearest word in the target language is the actual translation. Results are shown for each of the 4 embedding transformation combinations. Each distribution consists of 1000 observations (100 train-test splits for each of the 10 embedding spaces). 

P@1 results using cosine similarity and CSLS are reported in Table~\ref{tab_p@1}. CSLS outperformed cosine similarity in all conditions except for the test performance in the Korean-unaligned (PCA-256) condition. Therefore, we report CSLS results. When aligned, p@1 was close to 1 for words in the train dataset, but substantially dropped for words in the test dataset in both source languages. P@1 for the test words was better when unaligned than aligned, but was still worse than that of train words. These results replicate previous work on cross-lingual alignment, where p@1 for test words remained generally low, and p@1 often dropped after orthogonal transformation for pairs of typologically distinct languages \cite{peng2024concept}. For this reason, we include all four types of embedding transformations as features in constructing the best gap word classifier. See Figure~\ref{fig_appendix_pca_p@1} and Appendix~\ref{sec:appendix_pca_dim_p@1} for visualization and discussion on p@1 results across different PCA dimensionalities.

\begin{table}[t!]
\centering

\begin{subtable}{\columnwidth}
\centering
\caption{Train performance}
\label{tab:p1_train}

\resizebox{\columnwidth}{!}{
\begin{tabular}{llcc}
\hline
\textbf{Language} & \textbf{Transformation} & \textbf{CS} & \textbf{CSLS} \\
\hline

\multirow{4}{*}{English}
& Procrustes (All dims) & 0.999 & 0.999 \\
& Procrustes (PCA-256) & 0.990 & 0.992 \\
& Unaligned (All dims) & 0.404 & 0.423 \\
& Unaligned (PCA-256) & 0.356 & 0.394 \\
\hline

\multirow{4}{*}{Korean}
& Procrustes (All dims) & 0.991 & 0.992 \\
& Procrustes (PCA-256) & 0.973 & 0.975 \\
& Unaligned (All dims) & 0.409 & 0.427 \\
& Unaligned (PCA-256) & 0.330 & 0.377 \\
\hline
\end{tabular}
}

\end{subtable}

\vspace{0.8em}

\begin{subtable}{\columnwidth}
\centering
\caption{Test performance}
\label{tab:p1_test}

\resizebox{\columnwidth}{!}{
\begin{tabular}{llcc}
\hline
\textbf{Language} & \textbf{Transformation} & \textbf{CS} & \textbf{CSLS} \\
\hline

\multirow{4}{*}{English}
& Procrustes (All dims) & 0.154 & 0.176 \\
& Procrustes (PCA-256) & 0.183 & 0.204 \\
& Unaligned (All dims) & 0.393 & 0.415 \\
& Unaligned (PCA-256) & 0.338 & 0.345 \\
\hline

\multirow{4}{*}{Korean}
& Procrustes (All dims) & 0.137 & 0.157 \\
& Procrustes (PCA-256) & 0.174 & 0.187 \\
& Unaligned (All dims) & 0.397 & 0.417 \\
& Unaligned (PCA-256) & 0.361 & 0.357 \\
\hline
\end{tabular}
}

\end{subtable}

\caption{Precision@1 performance by embedding transformation combinations, source languages, and similarity metrics. Subtable (a) shows train performance, and subtable (b) shows test performance. `All dims' indicates using all embedding dimensions, while `PCA-256' indicates dimensionality-reduced embedding spaces with 256 dimensions.}
\label{tab_p@1}

\end{table}

\begin{table*}[p]
\centering
\setlength{\tabcolsep}{12pt}
\begin{tabular}{@{}l*{4}{c}@{}}
\toprule
Model-embedding Combination & $t$ & $d$ & $p$ & $p_{\text{FDR}}$ \\
\midrule
\textbf{Kanana-nano (last 4 layers)}            & 25.641 & \textbf{1.504} & $6.52 \times 10^{-111}$ & $1.24 \times 10^{-109}$ \\
\textbf{Exaone-3.5-7.8b (first layer)}       & 23.093 & \textbf{1.355} & $4.95 \times 10^{-92}$  & $4.20 \times 10^{-91}$  \\
\textbf{Kanana-nano (last layer)}             & 22.650 & \textbf{1.329} & $6.63 \times 10^{-92}$  & $4.20 \times 10^{-91}$  \\
\textbf{Kanana-nano (sentence embedding)}             & 22.108 & \textbf{1.297} & $2.92 \times 10^{-89}$  & $1.39 \times 10^{-88}$  \\
\textbf{Exaone-3.5-2.4b (last 4 layers)}       & 16.553 & \textbf{0.971} & $2.23 \times 10^{-54}$  & $8.47 \times 10^{-54}$  \\
\textbf{Exaone-3.5-2.4b (first layer)}       & 15.925 & \textbf{0.934} & $4.55 \times 10^{-51}$  & $1.44 \times 10^{-50}$  \\
\textbf{Kanana-1.5-8b (first layer)}         & 12.678 & \textbf{0.744} & $1.71 \times 10^{-34}$  & $4.65 \times 10^{-34}$  \\
\textbf{Kanana-1.5-2.1b (first layer)}       & 12.601 & \textbf{0.739} & $9.69 \times 10^{-34}$  & $2.30 \times 10^{-33}$  \\
\textbf{Exaone-3.5-7.8b (last 4 layers)}       & 12.429 & \textbf{0.729} & $2.47 \times 10^{-33}$  & $5.22 \times 10^{-33}$  \\
\textbf{Exaone-3.5-2.4b (last layer)}        & 10.574 & \textbf{0.620} & $6.16 \times 10^{-25}$  & $1.17 \times 10^{-24}$  \\
Kanana-nano (first layer)            & 10.473 & 0.614 & $1.90 \times 10^{-24}$  & $3.29 \times 10^{-24}$  \\
HCX-1.5b (last4 layers)               & 9.200  & 0.540 & $1.65 \times 10^{-19}$  & $2.61 \times 10^{-19}$  \\
Kanana-1.5-8b (last4 layers)         & 7.348  & 0.431 & $3.79 \times 10^{-13}$  & $5.54 \times 10^{-13}$  \\
Kanana-1.5-2.1b (last4 layers)       & 5.735  & 0.337 & $1.24 \times 10^{-8}$   & $1.68 \times 10^{-8}$   \\
Exaone-3.5-7.8b (last layer)        & 4.883  & 0.287 & $1.19 \times 10^{-6}$   & $1.51 \times 10^{-6}$   \\
HCX-1.5b (first layer)               & 4.575  & 0.268 & $5.28 \times 10^{-6}$   & $6.27 \times 10^{-6}$   \\
HCX-1.5b (last layer)                & 3.728  & 0.219 & $2.02 \times 10^{-4}$   & $2.26 \times 10^{-4}$   \\
Kanana-1.5-8b (last layer)           & 2.683  & 0.157 & $7.39 \times 10^{-3}$   & $7.80 \times 10^{-3}$   \\
Kanana-1.5-2.1b (last layer)         & 2.510  & 0.147 & $1.22 \times 10^{-2}$   & $1.22 \times 10^{-2}$   \\
\bottomrule
\end{tabular}
\caption{T-test results and effect sizes (Cohen's \textit{d}) on the differences in cosine similarity across matched and mismatched translation pairs. Model-embedding combinations are ordered by Cohen's \textit{d} values. `Last 4 layers' refers to the averaged embedding of the last four layers.}
\label{tab_appendix-stats}
\end{table*}

\section{Non-emotion Words as Additional Training Data}
\label{sec:appendix_non+emotion}

\subsection{Non-emotion words}
Korean and English non-emotion words were retrieved from the Korlex 1.5 dataset \cite{yoon2009construction}, a Korean WordNet constructed based on PWN (Princeton WordNet; \citealp{miller1995wordnet}). Unlike EVOKE, however, Korlex 1.5 does not explicitly annotate lexical gaps. Although some synsets included entries from only one language, they could not be reliably interpreted as lexical gaps without an extensive validation procedure.\footnote{A set of synonymous lexical items in English and Korean is organized into synsets, each assigned a unique identifier in Korlex 1.5. Each synset may contain a single lexical item, a multi-word expression, or a ``lexical blank'', where entries exist in only one language but not the other. Inspection on the dataset revealed that some ‘lexical blanks’ reflect incomplete entries rather than genuine untranslatability (e.g., although the word \textit{금고} or \textit{돈궤} could be easily translated as a \textit{safe} or \textit{vault}, their English counterparts are not listed in the dataset.) Addressing these cases would require extensive reconstruction of the dataset, which is beyond the scope of this paper. For this reason, we did not consider Korlex 1.5 as providing an explicit index of untranslatability, and therefore utilized it mainly for obtaining translatable, non-emotion adjectives.} To reduce the ambiguity and to obtain reliable experiments and evaluation, we restricted the non-emotion word set to items with a single lexical entry in the resource. These non-emotion words serve as an additional training source for the model to learn Korean-English lexical alignment.

\subsection{Machine Acceptabililty Judgments}
\label{sec:appendix_prompt}

\begin{table*}[t!]
\centering
\begin{tabular}{|p{0.45\textwidth}|p{0.45\textwidth}|}
\hline
\textbf{English Prompt} & \textbf{Korean Prompt} \\
\hline
You are judging sentence acceptability.

\textbf{Task:} Decide whether the sentence below is acceptable for a native English speaker, without additional context.

\textbf{Sentence:} ``I am \{word\}''

\textbf{Output format (STRICT):} Output only one character. 1 if acceptable, 0 if unacceptable. Do not explain.
&
당신은 문장 수용성을 판단합니다.

\textbf{과제:} 아래 문장이 추가 맥락 없이 한국어 화자에게 자연스럽고 수용 가능한지 판단하세요.

\textbf{문장:} ``나는 \{word\}''

\textbf{출력 형식 (엄격):} 반드시 한 글자만 출력하세요. 1이면 수용 가능, 0이면 수용 불가능. 설명하지 마세요. \\
\hline
\end{tabular}
\caption{Prompts for sentence acceptability judgment in English and Korean.}
\label{tab_prompts}
\end{table*}

For non-emotion words, we utilized two bilingual large language models, one small and one big model among the models selected as the top 10 model-embedding combinations, \texttt{kanana-1.5-8b-base} and \texttt{EXAONE-3.5-2.4B-Instruct} \cite{an2024exaone, bak2025kanana}, to filter candidate words that are acceptable in the same sentence frame used for emotion words. This was because Korlex 1.5 lacked human acceptability annotations. Only words judged acceptable in the sentence frame by both models were retained. 

Prompts in Korean and English (see Table~\ref{tab_prompts}) were used to obtain binary judgments (1 as acceptable and 0 as unacceptable) of words in each language from two models. To ensure compatibility across Korean and English prompts, the two prompts closely modeled each other in overall structure and word choice.

\begin{figure}[ht!]
    \centering
    \includegraphics[width=\columnwidth]{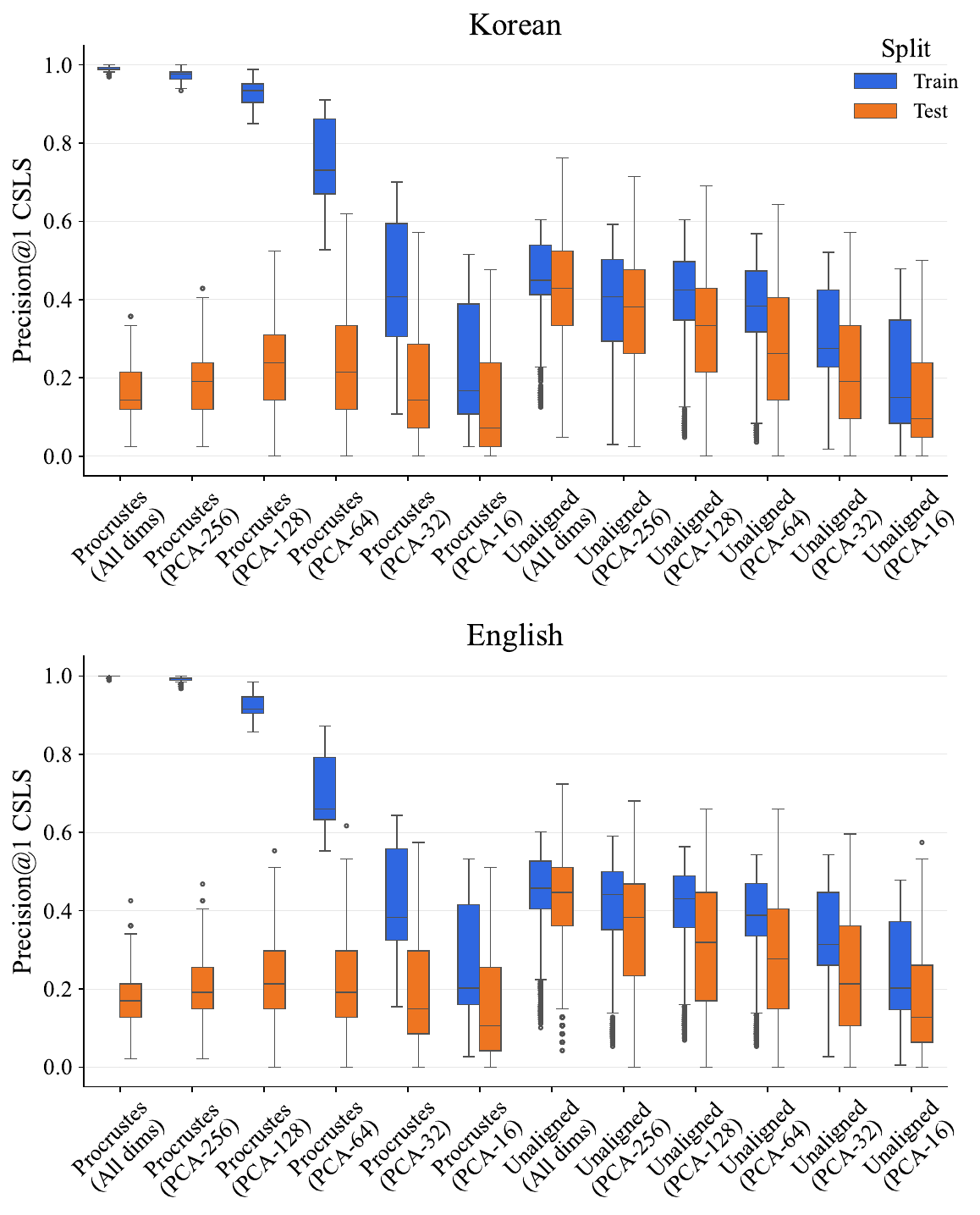}
    \caption{Distribution of train and test precision@1 performance by different PCA dimensionalities and alignment. `All dims' indicates using all dimensions of embeddings without dimensionality reduction. `PCA-N' indicates using dimensionality-reduced embedding spaces (N dimensions).}
    \label{fig_appendix_pca_p@1}
\end{figure}

\subsection{Gap vs Non-gap Test word AUC}
\begin{figure*}[p]
    \centering
    \includegraphics[width=\textwidth]{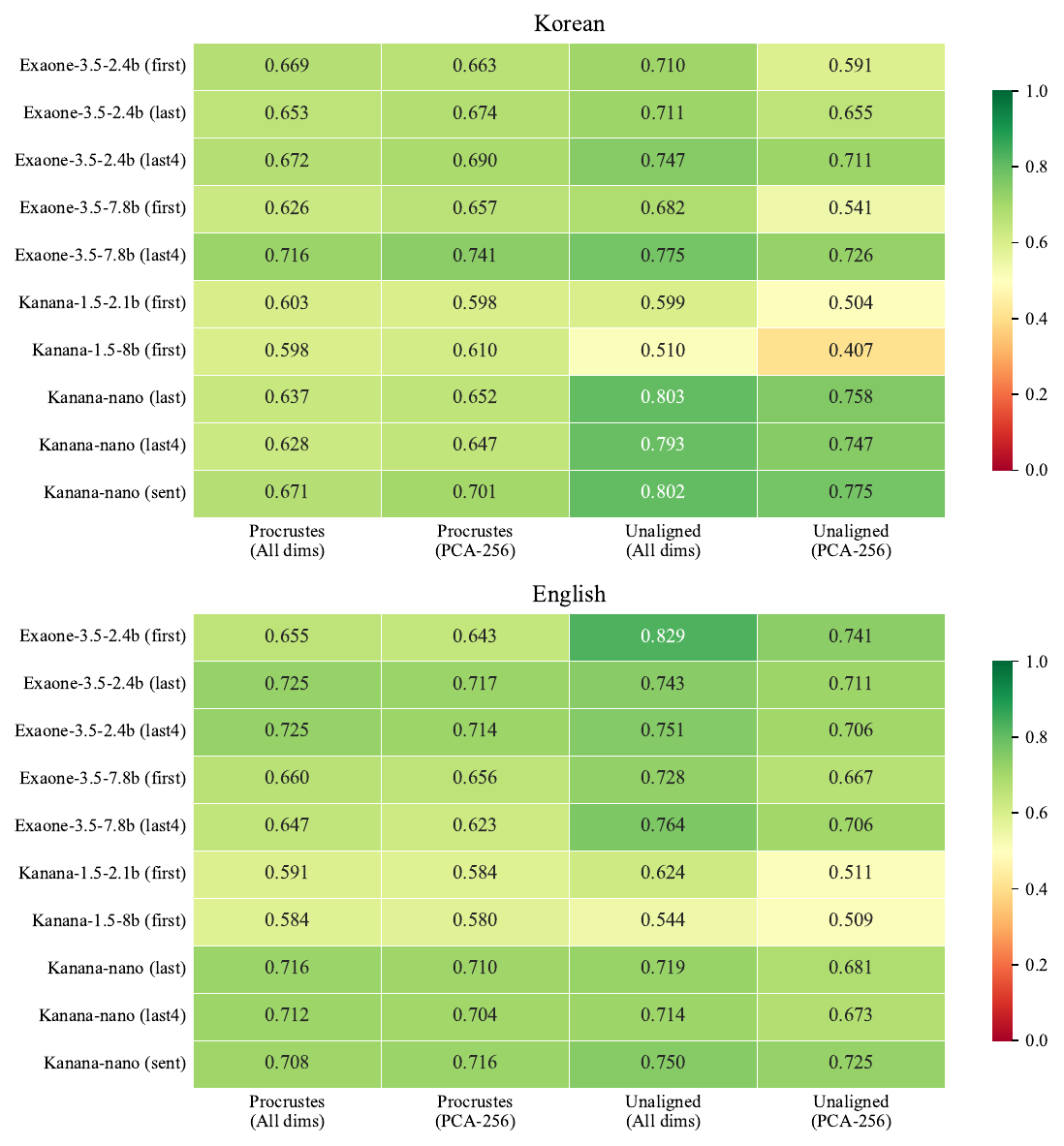}
    \caption{Gap vs Non-gap Test word AUCs across all languages and embedding spaces. The AUC score in each grid represents the median AUC across all 100 train-test splits for each embedding space. `Last 4' refers to the averaged embedding of the last four layers. `PCA-256' indicates using dimensionality-reduced embedding spaces (256 dimensions).}
    \label{fig_appendix_medianauc}
\end{figure*}

\begin{figure*}[p]
    \centering
    \includegraphics[width=\textwidth]{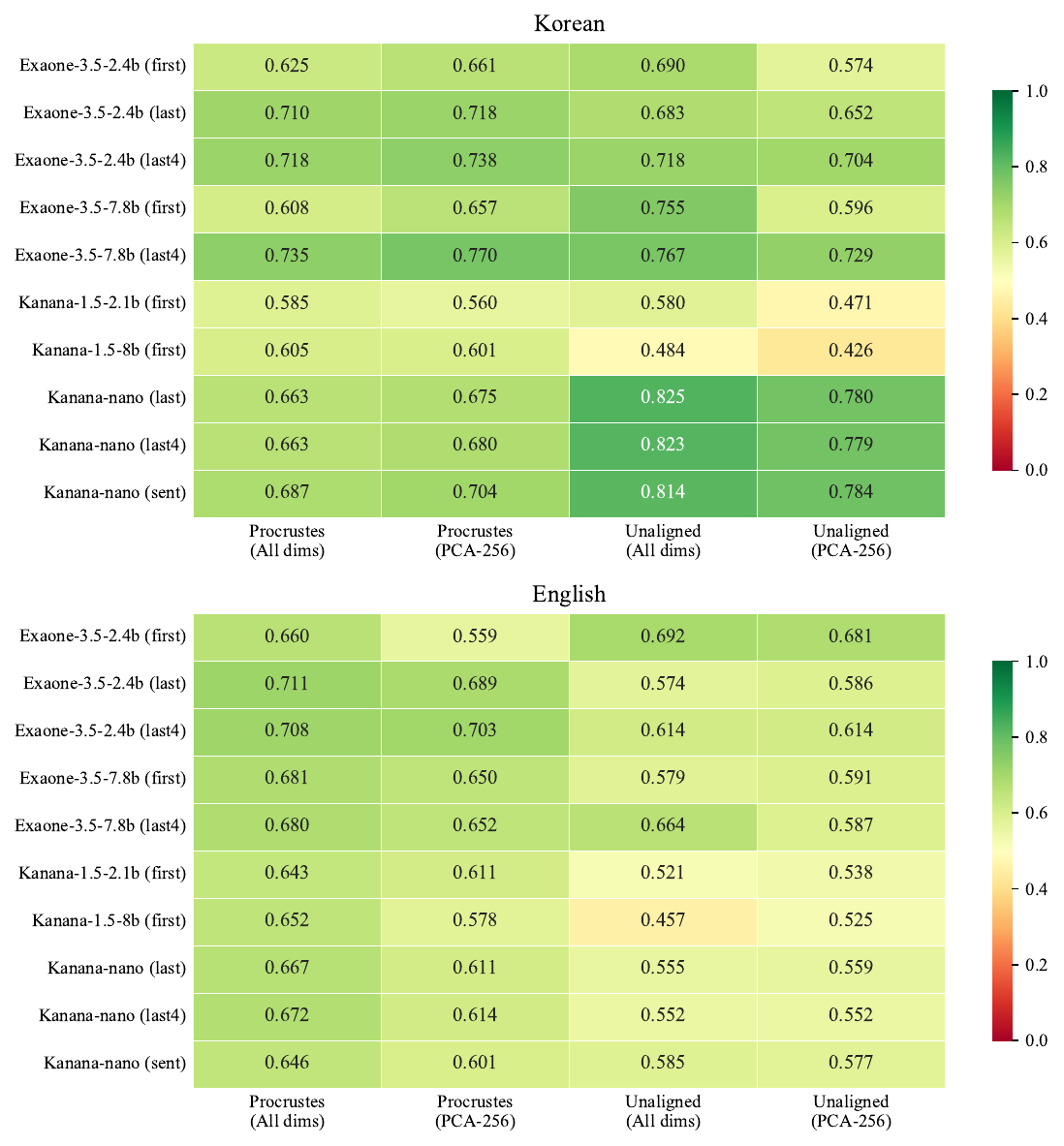}
    \caption{Gap vs Non-gap Test word AUCs of models trained using both emotion and non-emotion words across all languages and embedding spaces. The AUC score in each grid represents the median AUC across all 100 train-test splits for each embedding space. `Last 4' refers to the averaged embedding of the last four layers. `PCA-256' indicates using dimensionality-reduced embedding spaces (256 dimensions).}
    \label{fig_appendix_medianauc_more_train}
\end{figure*}

We evaluated the CSLS metric of embedding spaces derived from using both emotion and non-emotion Words for training. Figure~\ref{fig_appendix_medianauc_more_train} presents Gap vs Non-gap Test word AUC scores across embedding spaces (see Figure~\ref{fig_appendix_medianauc} for comparison with the emotion-only setting). Adding non-emotion words yielded a slight improvement in some of the embedding spaces for Korean, but overall performance for English was worse compared to the emotion-only setting.

One of the likely reasons could be the imbalance in translation mappings obtained from Korlex 1.5, where a smaller set of Korean words maps densely to a larger set of English words. As this study includes only one-to-one and many-to-one mappings for training and evaluation, this imbalance may have introduced cross-linguistic asymmetry.

Given that incorporating non-emotion adjectives did not substantially improve overall performance, and in fact substantially reduced performance for English, we did not proceed with training logistic classifiers on both emotion and non-emotion words. Overall, the limited and inconsistent cross-lingual benefits do not justify expanding the training set with additional non-emotion adjectives.

\begin{figure*}[p]
    \centering
    \includegraphics[width=\textwidth]{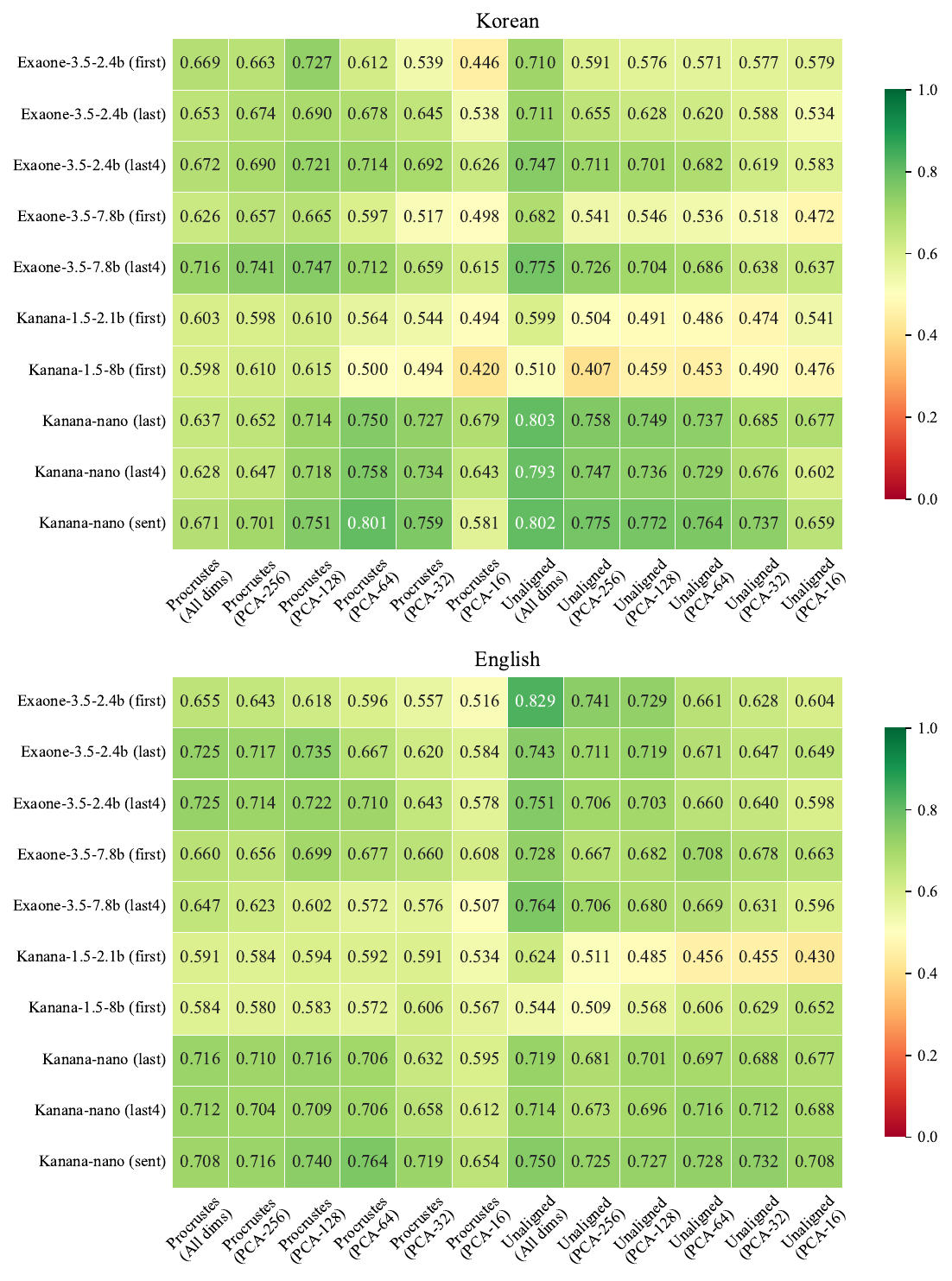}
    \caption{Gap vs Non-gap Test word AUCs across all languages and embedding spaces with varying PCA dimensionalities. The AUC score in each grid represents the median AUC across all 100 train-test splits for each embedding space. `Last 4' refers to the averaged embedding of the last four layers. `PCA-N' indicates using dimensionality-reduced embedding spaces (N dimensions).}
    \label{fig_appendix_pca_medianauc}
\end{figure*}

\begin{figure*}[hb!]
\centering

\begin{subtable}{0.48\textwidth}
\centering
\caption{All embedding spaces as predictors (N=40)}
\label{tab:l1_penalty_a}
\resizebox{\linewidth}{!}{
\begin{tabular}{lcccc}
\toprule
 & \multicolumn{2}{c}{AUC} & \multicolumn{2}{c}{N of retrieved gap words} \\
\cmidrule(lr){2-3} \cmidrule(lr){4-5}
$c$ & Korean & English & Korean & English \\
\midrule
0.05 & 0.82 & 0.76 & 18 & 25 \\
0.1  & 0.83 & 0.76 & 17 & 23 \\
0.2  & 0.82 & 0.74 & 14 & 22 \\
0.5  & 0.80 & 0.77 & 13 & 20 \\
1 & 0.80 & 0.74 & 11 & 17 \\
2  & 0.80 & 0.72 & 10 & 15 \\
5  & 0.78 & 0.72 & 9  & 13 \\
10 & 0.74 & 0.71 & 8  & 13 \\
\bottomrule
\end{tabular}}
\end{subtable}
\hfill
\begin{subtable}{0.48\textwidth}
\centering
\caption{Unaligned embedding spaces as predictors (N=20)}
\label{tab:l1_penalty_b}
\resizebox{\linewidth}{!}{
\begin{tabular}{lcccc}
\toprule
 & \multicolumn{2}{c}{AUC} & \multicolumn{2}{c}{N of retrieved gap words} \\
\cmidrule(lr){2-3} \cmidrule(lr){4-5}
$c$ & Korean & English & Korean & English \\
\midrule
0.05 & 0.75 & 0.75 & 18 & 26 \\
0.1  & 0.81 & 0.76 & 18 & 26 \\
0.2  & 0.79 & 0.75 & 15 & 23 \\
0.5  & 0.77 & 0.79 & 13 & 21 \\
1    & 0.78 & 0.78 & 13 & 19 \\
2    & 0.80 & 0.77 & 11 & 19 \\
5    & 0.79 & 0.76 & 11 & 19 \\
10   & 0.81 & 0.75 & 11 & 19 \\
\bottomrule
\end{tabular}}
\end{subtable}

\caption{Logistic regression results by L1 regularization parameter, $c$.}
\label{fig:appendix_l1_penalty}
\end{figure*}

\section{Experiments on PCA Dimensionalities}
\label{sec:appendix_pca_dim}

\subsection{Precision@1}
\label{sec:appendix_pca_dim_p@1}
We first explored p@1 across various PCA dimensionalities in both aligned and unaligned embedding spaces. We followed the procedure described in Section~\ref{sec:appendix_p@1}. Results are shown in Figure~\ref{fig_appendix_pca_p@1}. The overall trends were consistent across both source languages. In unaligned spaces, both the train and test p@1 generally decreased as dimensionality decreased. In contrast, aligned spaces exhibited a sharper decline in train performance as the dimensionality decreased, while the test performance remained relatively better at dimensionalities of 64 and 128. This suggests that a mid-range dimensionality may provide modest benefits in aligned spaces. Nevertheless, overall test performance was consistently highest in unaligned high-dimensional spaces.

\subsection{Gap vs Non-gap Test word AUC}
We next examined the distributions of cross-lingual nearest neighbor CSLS scores across all embedding spaces with a wider range of PCA dimensionalities. Results are reported in Figure~\ref{fig_appendix_pca_medianauc}. In both source languages, the best-performing embedding configuration was the unaligned space without dimensionality reduction. Among the aligned embedding spaces, dimensionality of 64 yielded the best AUC for both languages. Overall, these findings are consistent with the p@1 results: unaligned high-dimensional spaces produced the strongest performance, while mid-range dimensionality yielded the best performance within aligned spaces.

\subsection{Logistic Classifier Performance}
Lastly, we examined the logistic classifier performance using different PCA dimensionality for unaligned and aligned spaces. For embedding configurations using PCA to reduce dimensionality, we tested 64 dimensions for unaligned spaces and 256 for aligned spaces. While the classifier performance with an L1 penalty ($C$ = 0.1) was better in full models (using all embedding spaces) compared to reduced models (using only the unaligned spaces) in AUC (Korean; AUC = 0.85 versus AUC = 0.81, English; AUC = 0.81 versus AUC = 0.76), the retrieval was better in reduced models (Korean; 18/19 versus 18/19 retrieval, English; 23/27 versus 26/27 retrieval). The results suggest that classifiers using aligned spaces can be optimized through testing multiple PCA dimensionalities. However, a trade-off was observed between AUC and retrieval score, while unaligned spaces continue to yield good overall performance. 

\clearpage
\section{Application}
\label{sec:appendix_application}
Below are the steps one could take to apply the proposed framework for identifying lexical gaps from new lexical resources. These recommendations are based on the simplest possible procedure, motivated by our finding that lexical gap classifiers trained on embeddings from bilingual MLLMs reliably capturing translational equivalence yield good performance without further alignment.

\begin{enumerate}
  \item First, prepare separate monolingual resources or multilingual resources with comprehensive lexical coverage in each language within the same domain.
  \item Second, select the MLLMs and extract different layers (first, last, the average of the last 4 layers) or sentence embeddings (if available) from each MLLM. If multiple MLLMs are available, and if the translation mappings between words in different languages are available, select the top-performing model-embedding combinations based on how well each embedding from a specific MLLM reliably captures translational equivalence. 
  \item Third, obtain the embeddings of each word. If the translation mappings are available, select only 1 target word per source word to ensure clean mappings. 
  \item Fourth, for each model-embedding combination, go through train-test splits and perform dimensionality reduction (e.g., PCA) using the embeddings from two languages (source and target) in the train set.
  \item In each of the resulting embedding spaces (e.g., 100 random seeds for train-test splits × 10 model-embedding combinations × 2 embedding transformations), compute the CSLS score between each source word and its nearest neighbor in the target language.
  \item Build logistic classifiers (with LOOCV) using each target word's median CSLS scores across random seeds per each embedding configuration as predictors.
  \item Using the estimated probability (\textit{gappiness} threshold), retrieve words with an estimated probability of 0.5 or higher.
  \item Recruit bilingual experts speaking both the source and target languages to identify lexical gaps among the retrieved word sets. Only the words with probabilities over 0.5 need to go through the additional manual inspection. Polysemous or compound words require special targeted inspection.
\end{enumerate}

\end{document}